\documentclass[10pt,twocolumn,letterpaper]{article}

\usepackage[pagenumbers]{cvpr} 
\usepackage{algorithm}
\usepackage[accsupp]{axessibility}
\usepackage{algorithmic}
\usepackage{subcaption}
\usepackage{booktabs}
\usepackage{amsmath}
\usepackage{multirow}
\usepackage{xcolor}
\usepackage{amssymb}
\usepackage{verbatim}

\usepackage{bbding}
\usepackage{makecell}
\usepackage{soul}
\usepackage{colortbl}
\definecolor{lightgray}{gray}{.9}

\definecolor{cvprblue}{rgb}{0.21,0.49,0.74}
\usepackage[breaklinks,colorlinks,allcolors=cvprblue]{hyperref}

\def\paperID{1029} 
\def\confName{CVPR}
\def\confYear{2026}

\title{RiskProp: Collision-Anchored Self-Supervised Risk Propagation for Early Accident Anticipation}

\author{
Yiyang Zou$^{1}$\thanks{Equal contribution.} \quad
Tianhao Zhao$^{1,2}$\footnotemark[1] \quad
Peilun Xiao$^{3}$ \quad
Hongyu Jin$^{3}$ \quad
Longyu Qi$^{3}$ \quad
Yuxuan Li$^{3}$ \\
Liyin Liang$^{3}$ \quad
Yifeng Qian$^{3}$ \quad
Chunbo Lai$^{3}$ \quad
Yutian Lin$^{1}$\thanks{Corresponding author.}  \quad
Zhihui Li$^{4}$ \quad
Yu Wu$^{1}$\footnotemark[2] \\
$^{1}$School of Computer Science, Wuhan University \quad
$^{2}$Zhongguancun Academy, Beijing, China \\
$^{3}$Didi Chuxing \quad
$^{4}$University of Science and Technology of China \\
\normalsize{\texttt{\{yiyangzou, happytianhao, yutian.lin, wuyucs\}@whu.edu.cn}}
}

\begin{document}
\maketitle
\begin{abstract}
Accident anticipation aims to predict impending collisions from dashcam videos and trigger early alerts. Existing methods rely on binary supervision with manually annotated “anomaly onset” frames, which are subjective and inconsistent, leading to inaccurate risk estimation. In contrast, we propose \textbf{RiskProp}, a novel collision-anchored self-supervised risk propagation paradigm for early accident anticipation, which removes the need for anomaly onset annotations and leverages only the reliably annotated collision frame. RiskProp models temporal risk evolution through two observation-driven losses: first, since future frames contain more definitive evidence of an impending accident, we introduce a future-frame regularization loss that uses the model’s next-frame prediction as a soft target to supervise the current frame, enabling backward propagation of risk signals; second, inspired by the empirical trend of rising risk before accidents, we design an adaptive monotonic constraint to encourage a non-decreasing progression over time. Experiments on CAP and Nexar demonstrate that RiskProp achieves state-of-the-art performance and produces smoother, more discriminative risk curves, improving both early anticipation and interpretability. Code: \href{https://github.com/xingyueye5/RiskProp/}{https://github.com/xingyueye5/RiskProp/}.
\end{abstract}
    
\section{Introduction}

As autonomous vehicles and advanced driver assistance systems continue to evolve, accident anticipation has become increasingly important for improving road safety. Its goal is to predict whether an accident is likely to occur in the near future based on the current driving scene captured by a dashcam. To achieve this, the system continuously estimates a risk score in real time, and if the score exceeds a predefined threshold, an alert is triggered to enable early intervention. This allows both autonomous systems and human drivers to take preventive actions before an accident happens, thereby reducing the risk of collision and enhancing overall driving safety.

\begin{figure}[t]
     \centering
         \subfloat[Previous binary frame-level risk labeling strategy for accident videos]{\includegraphics[width=\linewidth]{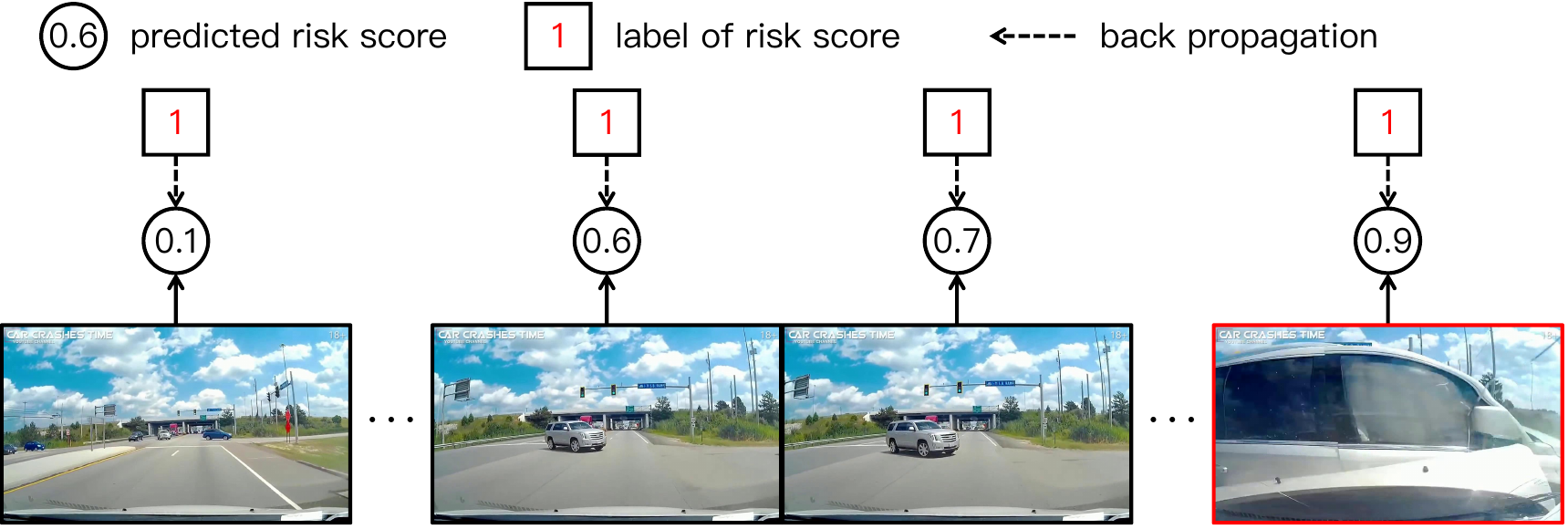}}
        \vspace{0.2cm} 
         \subfloat[Self-supervised frame-level risk propagation for accident videos]{\includegraphics[width=\linewidth]{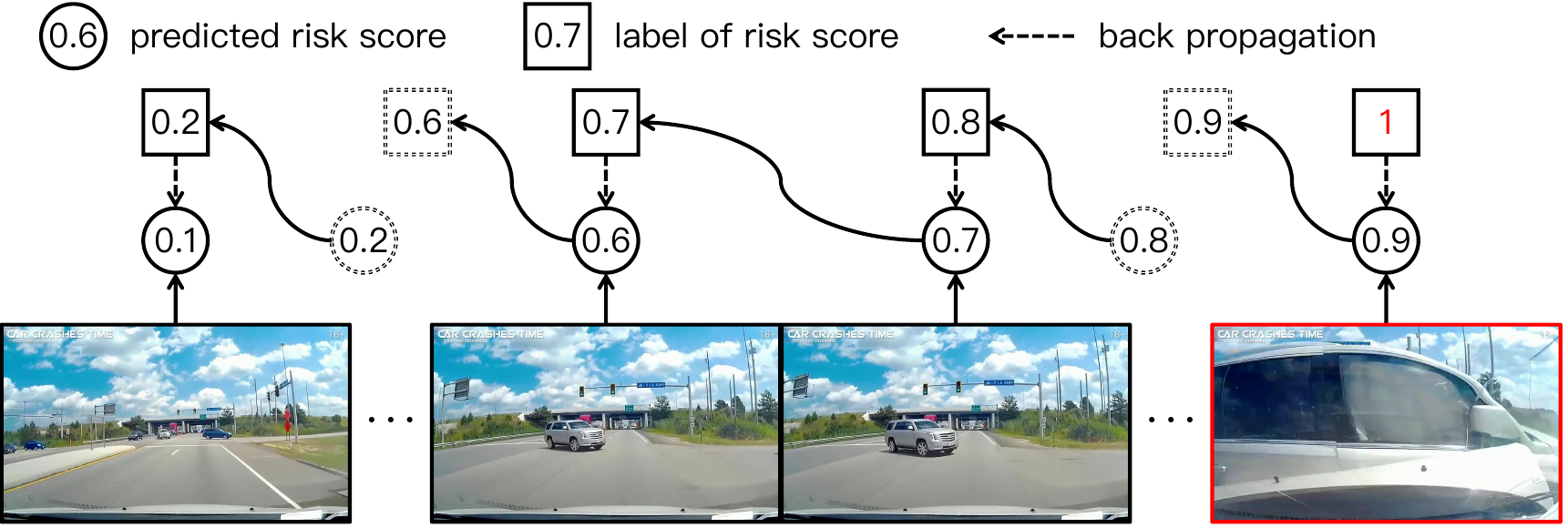}}
     \caption{Form of supervision of previous works and ours. (a) Most previous methods treat all frames in accident-free videos as negative samples with label 0, and treat frames in accident videos—from the annotated anomaly onset frame to the collision frame—as positive samples with labeled 1. (b) Our method treats the model’s prediction for the next frame as a supervision signal for the current frame, which enables the risk values to propagate gradually backward from the collision frame.
     }
     \label{fig:figure1}
\end{figure}

Most existing works~\cite{DAD, karim2022dynamic, liao2024real, Zhang2025LATTE,wu2020learning} for accident anticipation train models to predict a risk score for each frame by treating the task as supervised learning with binary labels: all frames in accident-free videos are labeled 0, while all frames in accident videos—from the beginning up to collision—are labeled 1, as shown in Fig.~\ref{fig:figure1} (a). To compensate for the fact that risk in pre-collision frames is typically intermediate and evolves gradually, these methods apply exponentially decaying weights to the binary cross-entropy loss, assigning lower penalties to earlier frames. To further improve early anticipation, some works~\cite{Adaptive-Loss, RARE} adopt adaptive loss weighting schemes, while others~\cite{CAP, nexar2025dashcamcollisionprediction} manually annotate the “anomaly onset” frame for accident videos and adjust loss weights relative to it. However, this binary labeling paradigm is fundamentally flawed: it forces the model to treat all pre-collision frames as equally risky, ignoring the intermediate and scenario-dependent nature of risk evolution—such as slow increases when a driver is distracted \textit{versus} rapid spikes when a pedestrian suddenly appears. Moreover, the manual annotation of anomaly onset is subjective and inconsistent across annotators, leading to noisy and unreliable supervision. As a result, the model is misled to optimize risk predictions toward inaccurate binary targets rather than reflecting the true, dynamic progression of risk.

Given the limitations of handcrafted labels and unreliable manual annotations, we turn to data-driven observations about the nature of risk in driving scenarios. First, we observe that accident anticipation fundamentally aims to predict whether a collision will occur in the future based on current observations. Since future frames contain more definitive evidence about the occurrence of an accident, the model’s risk predictions for these frames are typically more accurate and better aligned with ground truth than those for earlier frames. This suggests that future predictions can serve as reliable pseudo-supervision signals to guide the learning of current risk estimates. Second, we find that in accident videos, the underlying risk tends to follow a non-decreasing trend over time within a long temporal window before collision—reflecting the progressive deterioration of driving safety as hazardous conditions unfold. 

Building on these observations, we propose RiskProp, a novel collision-anchored self-supervised risk propagation paradigm for early accident anticipation. It generates intermediate risk scores for pre-collision frames by propagating supervision backward from the collision frame without relying on binary labeling schemes or manual anomaly onset annotations. 
Specifically, as shown in Fig.~\ref{fig:figure1} (b), for accident videos, we assign a hard risk label of 1 only to the collision frame, whose timing is reliably annotated. For all other pre-collision frames in accident videos, we generate soft supervision signals by propagating risk information backward in time instead of fixed binary labels, where the model’s prediction on the next frame (detached from gradient computation) serves as the target for the current frame. This self-supervised temporal regularization enables early frames to receive intermediate and discriminative risk supervision, reflecting the gradual buildup of danger rather than being forced into arbitrary hard labels. For accident-free videos, all frames are assigned a risk label of 0. By design, RiskProp avoids the need for subjective anomaly onset annotation and learns a more nuanced risk progression grounded in temporal consistency. Furthermore, to encourage a plausible long-term trend in risk evolution, we introduce an adaptive monotonic constraint loss that penalizes violations where a later frame is assigned a lower risk than an earlier one within the same sequence. This loss operates over randomly sampled frame pairs and allows short-term fluctuations while promoting an overall non-decreasing pattern consistent with real-world dynamics.

Our contributions can be summarized as follows:  
\begin{itemize}
    \item We reveal the limitations of binary labeling and manual anomaly onset annotations in accident anticipation, and instead base our approach on two empirical observations: future frames provide more reliable risk cues, and pre-collision risk tends to grow non-decreasingly over time.

    \item We propose RiskProp, a novel collision-anchored self-supervised paradigm that generates intermediate risk scores by propagating supervision backward from the collision frame. It introduces a future-frame regularization loss and an adaptive monotonic constraint loss, eliminating the need for subjective annotations and encouraging early anticipation.
    
    \item Extensive experiments on two datasets demonstrate that our method achieves state-of-the-art performance and produces risk curves with better temporal consistency and earlier warning capability.
\end{itemize}

\section{Related Work}

\begin{figure*}[t]
    \centering
    \includegraphics[width=\textwidth]{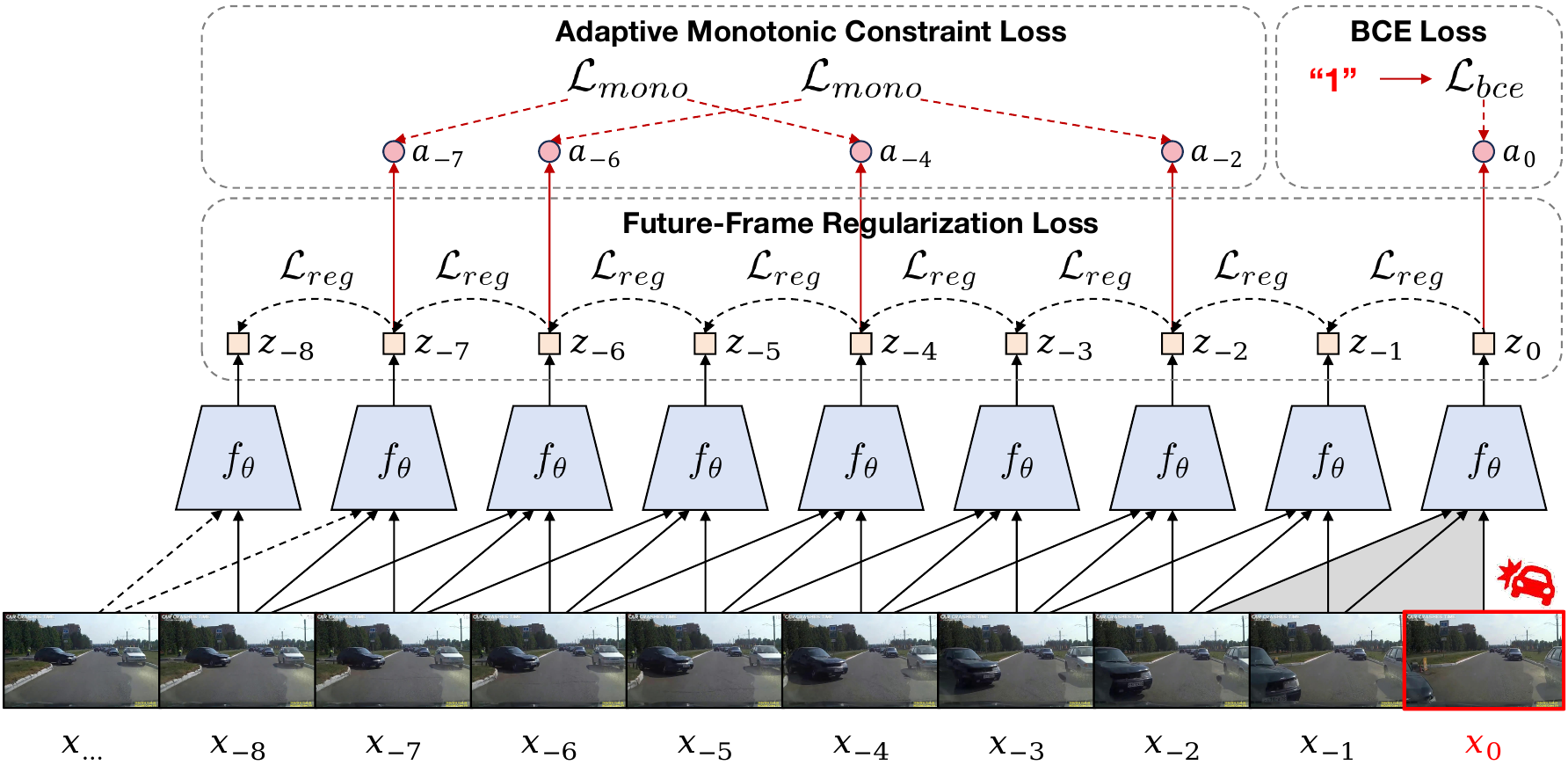}
    \caption{\textbf{Overview of our RiskProp framework.} The encoder-only model takes a snippet of consecutive frames and predicts the current frame's risk score. 
    To train such a model without ``anomaly onset'' labels and with only the objective collision frame label, two losses are proposed: 
    The \textbf{Future-Frame Regularization Loss} uses the next-frame's detached prediction as a self-supervised target for high-risk signals backward propagation, while the \textbf{Adaptive Monotonic Constraint Loss} imposes a monotonicity constraint to ensure a non-decreasing overall risk trend. And we adopt the Binary Cross-Entropy loss to provide explicit supervision only at the collision frame. This enables stable, physically plausible risk curves without manual onset annotations.}
    \label{fig:overview}
\end{figure*}

\subsection{Traffic Accident Anticipation}

Existing accident anticipation methods predominantly adopt a frame-level binary supervision paradigm that predicts a risk score for each frame. In terms of label supervision, some works ~\cite{DAD, Adaptive-Loss, CCD} define positive samples using a fixed temporal window before collision, while others ~\cite{Dada-2000,CAP} rely on manually annotated anomaly onset label. Several approaches further introduce collision-time-anchored exponential or adaptive loss weighting schemes~\cite{XAI,DAD}, which emphasize frames closer to the accident in training. However, both rigid temporal windows and subjective onset annotations introduce strong assumptions about risk evolution, often resulting in ambiguous supervision and temporally volatile or physically implausible risk predictions.

Beyond supervision design, many works leverage auxiliary cues such as driver gaze~\cite{DADA} or causal reasoning ~\cite{MM-AU} to improve interpretability, while others employ object detectors or attention maps for better spatial grounding ~\cite{RARE,li2023traffic,CAP,karim2022dynamic,zhao2024improving}. Transformer-based architectures have also been widely adopted to capture long-range temporal dependencies in accident anticipation~\cite{AVT, MeMViT,kumamoto2025aat,anik2024time,al2024traffic,ye2022vptr,roy2024interaction,guan2023egocentric}. Specifically, Wang et al.~\cite{wang2023memoryand} jointly models long-term memory and short-term anticipation to achieve fast anticipation, while several works~\cite{HRO,ego,wang2023gsc} model egocentric action prediction using graph structures. RAFTformer~\cite{RAFTformer} further introduces a future-aware teacher–student design for real-time action forecasting.
Despite these architectural advances, most existing methods remain fundamentally constrained by rigid or subjective binary supervision schemes.

\subsection{Self-Supervised Temporal Modeling in Video}

Self-supervised learning has emerged as a powerful paradigm for leveraging temporal structure in videos. Early methods exploit temporal order~\cite{misra2016shuffle} or cycle consistency~\cite{dwibedi2019temporal} as pretext tasks to learn video representations. Contrastive learning frameworks~\cite{sermanet2018time,han2021video} pull nearby frames closer in embedding space, enforcing local smoothness.  Some works ~\cite{yeche2023temporal,camporese2021knowledge} use label smoothing for early event prediction. A line of work uses future-aware ``teacher" models to guide real-time ``student" models~\cite{RAFTformer,tran2021knowledge}, enabling anticipation through knowledge distillation. Ristea et al.~\cite{ristea2024selfdistill} propose a self-distilled masked auto-encoder framework for effective video anomaly detection.
Our approach is conceptually related but significantly simpler: we use the next frame’s prediction from the same model (with gradient detachment) as a self-supervised target. This design avoids architectural complexity while effectively propagating high-risk signals backward in time.

\subsection{Monotonicity in Sequential Prediction}
Modeling irreversible processes often requires monotonic behavior~\cite{chen2023address}. In accident anticipation, several works enforce increasing risk trends via weighted losses~\cite{DAD} or adaptive penalties~\cite{Adaptive-Loss}. Pairwise ranking losses~\cite{CCD} further encourage later frames to have higher scores than earlier ones, promoting local monotonicity. Some incorporate uncertainty modeling to stabilize predictions~\cite{mhammedi2021risk}, but still operate within the supervised annotation paradigm and rely on manually defined labels. By enforcing a non-decreasing danger assumption, \cite{eccv2024self} imposes temporal monotonicity but fails to capture false-alarm scenarios in which perceived risk rises despite no actual accident.
Our method enforces a non-decreasing long-term risk trend while allowing short-term fluctuations, yielding smooth and physically plausible risk curves aligned with real driving dynamics.

\section{Method}
\subsection{Framework Overview}
\label{sec:temporal_anticipation}

Given a dashcam video sequence, the goal of traffic accident anticipation is to estimate, at each time step, the probability that a collision will occur in the near future. 
The model aims to encode the past and current frames and outputs the risk score for the current frame. An accident alert is triggered if this score exceeds a predefined threshold. 

In this paper, we propose a novel RiskProp framework, which models the temporal evolution of risk by leveraging both objective supervision from the collision frame and self-supervised temporal consistency. As illustrated in Fig.~\ref{fig:overview}, for the timestamp $t$, the input frames are denoted as $\mathbf{x}_t= \{x_{t-O+1}, \dots, x_{t} \}$, where $O$ is the number of observed frames ($O=3$ in the figure). The model encodes this snippet and outputs a risk representation $z_t=f_\theta(\mathbf{x}_t)$, which is then transformed into a risk score through a sigmoid activation, $a_t = \sigma(z_t)$ ($a_t \in (0,1)$). 

Our training objective is designed to guide these risk predictions towards a realistic and temporally consistent evolution pattern. As shown in Fig.~\ref{fig:overview}, the BCE loss and two auxiliary losses are adopted: (1) the future-frame regulation loss propagates risk signals backward from the collision frame; (2) the adaptive monotonic constraint loss encourages roughly non-decreasing risk curves. Together, these components enable RiskProp to learn risk curves that closely approximate real-world risk evolution. 

\subsection{Future-Frame Regularization Loss}
A key characteristic of driving-risk evolution is its temporal continuity: hazardous cues emerge gradually, and future frames usually provide stronger and more decisive evidence of an upcoming accident than earlier ones.
This suggests a natural self-supervision signal: the predicted risk score of the next frame can serve as a soft label for the current frame, enabling backward propagation of high-risk signals from the collision frame to earlier timesteps.

To leverage this characteristic, we introduce the \textbf{Future-Frame Regularization Loss}. Specifically, let $ \texttt{detach}(\cdot) $ denote the stop-gradient operation, which prevents gradients from flowing back through $ f_\theta(\mathbf{x}_{t+1}) $ during backpropagation. 
We define our future-frame regularization loss as:
\begin{equation}
\label{eq:reg_loss}
\mathcal{L}_{\text{reg}} = \sum_{t=1}^{T-1} \left\| \texttt{detach}(z_{t+1})- z_{t}\right\|^2,
\end{equation}
where the collision occurs at time $T$. The operation treats the detached variable $ \texttt{detach}(z_t) $ as a \emph{frozen target} rather than a trainable parameter. 
Moreover, this loss establishes a backward credit assignment pathway. Since the collision frame $ T $ has the only ground-truth label $ y_T = 1 $, its high predicted score propagates backward through the chain of $ \mathcal{L}_{\text{reg}} $ terms:
\[
\texttt{detach}(z_{T+1})  \to z_{T} ,\quad
\texttt{detach}(z_{T}) \to z_{T-1},\quad \dots
\]
Thus, even without explicit onset labels, early frames receive indirect supervision from the definitive hazard signal at the collision point. Through this backward propagation of risk information, the model learns to recognize early-stage cues that precede accidents, such as subtle lane drifts, delayed reactions, or unsafe following distances.

\begin{figure}[t]
    \centering
    \includegraphics[width=0.45\textwidth]{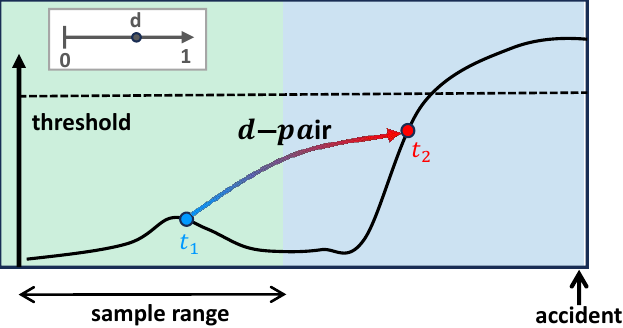}
    \caption{Illustration of sampling strategy for the adaptive monotonic constraint loss. For a randomly selected $d \in [d_{min}, d_{max}]$, a starting frame $t_1$ is sampled from $[0, T(1 - d)]$, and $t_2 = t_1 + dT$. This encourages learning across various temporal distances.}
    \label{fig:slide}
\end{figure}

\subsection{Adaptive Monotonic Constraint Loss}

In real-world accident scenarios, the level of danger generally increases as collision approaches. To capture this inherent progression, we introduce an \textbf{Adaptive Monotonic Constraint Loss} that encourages the model’s predicted risk scores to exhibit an overall non-decreasing trend along the temporal axis. This enforces a physically plausible evolution of risk while maintaining flexibility for minor fluctuations.

Formally, for sampled frame pairs $(x_i, x_j)$ where $j>i$, we expect:
\begin{equation}
    a_j \ge a_i.
\end{equation}

To leverage this hypothesis, we define a differentiable monotonic constraint loss to penalize cases where a later frame is predicted to be less risky than an earlier frame:
\begin{equation}
\mathcal{L}_{\text{mono}} 
= \frac{1}{|\mathcal{D}|} 
\sum_{(i,j) \in \mathcal{D}} 
\max \big( 0,\, a_i - a_j + \delta(\Delta t, \bar{c}_{i:j}) \big),
\end{equation}
where $\mathcal{D}$ denotes the set of randomly sampled frame pairs, and $\delta(\Delta t, \bar{c}_{i:j})$ is an adaptive tolerance margin controlling how strictly monotonicity is enforced. 

The margin $\delta$ adapts from two perspectives: (1) The temporal distance $\Delta t$. The margin increases with the temporal distance $\Delta t = t_j - t_i$, reflecting the expectation that risk may rise over longer intervals. (2) The confidence score $\bar{c}_{i:j}$. The enforcement strength depends on how confident the model is in its risk estimates. When the predicted risk scores are close to 0 or 1 (high confidence), the constraint should be stricter; when they are near the mean (low confidence), the constraint should relax. Formally, the adaptive tolerance margin is defined as:
\begin{equation}
\delta(\Delta t, \bar{c}_{i:j}) = \delta_0 \cdot \Delta t \cdot \bar{c}_{i:j},
\end{equation}
where $\delta_0$ is a scaling coefficient, $\Delta t$ measures the relative distance between two frames, and $\bar{c}_{i:j}$ denotes their average prediction confidence. For convenience, $\bar{c}_{i:j}$ is computed as:
\begin{equation}
\bar{c}_{i:j} = \frac{c_i + c_j}{2},
\quad
c_i = 2|a_i - \bar{a}|,c_j = 2|a_j - \bar{a}|,
\label{eq:margin}
\end{equation}
where $\bar{a}$ denotes the batch-wise mean risk score computed over positive videos in the batch, reflecting the typical risk levels in driving scenarios. A prediction farther from $\bar{a}$ is treated as more confident because it lies closer to either low-risk or high-risk extremes, while predictions near $\bar{a}$ indicate ambiguous intermediate states. This formulation ensures that higher confidence leads to stronger monotonic constraints.

This adaptive formulation balances flexibility and stability by adjusting the monotonic constraint to temporal distance and prediction confidence. It relaxes the constraint for short intervals or uncertain predictions, preventing over-regularization, while tightening it when frames are far apart or confidence is high, ensuring a globally consistent rising trend of risk over time.

\textbf{Sample strategy.} As illustrated in Fig.~\ref{fig:slide}, to ensure temporal diversity, we randomly sample frame pairs using offsets $d \in [d_{\min}, d_{\max}]$. Given $d$, we then sample a starting frame $i$ uniformly at random from the interval $[0, T(1-d)]$, where $T$ is the total length of the video clip, and set $j = i + dT$.

This sampling strategy offers two benefits. By randomly varying the temporal offset $d$, the model learns from both short- and long-term dynamics, improving robustness across different accident stages. Moreover, it aligns with the physical continuity of driving: while risk may rise gradually at first, it should not decrease once danger unfolds. The loss thus regularizes the risk curve to remain non-decreasing in expectation, yielding smooth and realistic progression.

\begin{table*}[t]
\caption{Quantitative results comparison of different methods on the CAP~\cite{CAP} dataset and the Nexar~\cite{nexar2025dashcamcollisionprediction} dataset.}
\centering
\begin{tabular}{l|cccc|cccc}
\toprule
\multirow{2}{*}{Method} & \multicolumn{4}{c|}{CAP~\cite{CAP}} & \multicolumn{4}{c}{Nexar~\cite{nexar2025dashcamcollisionprediction}} \\
\cmidrule{2-9}
& mAUC$^{0.1}$ & mAUC & mAP & mTTA$^{0.1}$ (s) & mAUC$^{0.1}$ & mAUC & mAP & mTTA$^{0.1}$ (s) \\
\midrule
AdaLEA~\cite{Adaptive-Loss} & 0.379 & 0.807 & 0.857 & 1.115 & 0.378 & 0.828 & 0.832 & 0.858 \\
XAI~\cite{XAI} & 0.352& 0.821 & 0.864 & 1.082 & 0.346 & 0.824 & 0.802  &0.839 \\
DSTA~\cite{karim2022dynamic} & 0.361 & \textbf{0.895} & 0.882 & 0.894 & 0.242 & 0.783 & 0.764 & 0.493 \\
GSC~\cite{wang2023gsc} & 0.378 & 0.881 & \textbf{0.890} & 0.911 & 0.322 & 0.802 & 0.811 & 0.815 \\
CAP~\cite{CAP} & 0.332 & 0.811 & 0.852 & 0.933 & 0.315 & 0.817 & 0.793 & 0.801 \\
CRASH~\cite{CRASH} &0.401 &0.842 &0.887 & 1.085 & 0.393 & 0.832 & 0.846 & 0.857\\
\textbf{Ours} & \textbf{0.483} & 0.853 & \textbf{0.890} & \textbf{1.207} & \textbf{0.472} & \textbf{0.869} & \textbf{0.870} & \textbf{0.958} \\
\bottomrule
\end{tabular}
\label{tab:comparison}
\end{table*}

\subsection{Labeling Strategy and Training Objective}
\label{subsec:labeling}

A major challenge in accident anticipation lies in the absence of reliable supervision. Manual annotations of “anomaly onset” are often subjective and inconsistent, whereas the collision timestamp provides an unambiguous and physically grounded reference. 
Accordingly, we adopt a simple labeling strategy: the actual collision frame is labeled as positive (label = 1), while only the initial start frame that is far from the accident is labeled as negative (label = 0). All intermediate frames are left unlabeled, and their risk values are inferred through self-supervised risk propagation.

We adopt a weighted Binary Cross-Entropy (BCE) loss. For accident videos, the loss is applied only to the starting frame and the collision frame:
\begin{equation}
\mathcal{L}_{\text{bce}} = -\frac{1}{|\mathcal{S}|} \sum_{i \in \mathcal{S}} w_i\left[y_i \log a_i + (1 - y_i) \log (1 - a_i) \right],
\end{equation}
where $\mathcal{S}$ denotes the set of two frames that are either the first frame of a training snippet or the accident frame, $a_i$ is the model's predicted danger score for frame $i$, $y_i \in \{0,1\}$ is the label, and \(w_i\) is the frame weight. To alleviate the severe supervision imbalance between sparse positive anchors and abundant negative frames, we assign a higher weight to the collision frames than to negative frames.
For non-accident videos, since no collision anchor is available and the monotonicity assumption does not hold, both the future-frame regularization and the adaptive monotonic constraint are disabled. Therefore, we supervise these sequences by assigning all frames a negative label ($y_t=0$ for all $t$) and optimizing the BCE loss over the entire video. The full training objective is computed as:
\begin{equation}
\label{eq:total_loss}
\mathcal{L} = \mathcal{L}_{\text{bce}} + \lambda_1 \cdot \mathcal{L}_{\text{reg}} + \lambda_2 \cdot \mathcal{L}_{\text{mono}},
\end{equation}
where $\lambda_1$ and $\lambda_2$ are hyperparameters balancing the contributions of the regularization terms. This formulation ensures that the model learns to anticipate danger not by memorizing subjective labels, but by discovering the underlying temporal structure of risk evolution, with the collision frame acting as the sole anchor point for supervision.

\section{Experiments}
\subsection{Experimental Setup}
\textbf{Datasets.} We validated the effectiveness of our method using experiments on two real-world traffic accident datasets: \textbf{CAP}~\cite{CAP} comprises 11,727 ego-view accident videos, totaling 2,195,613 frames, and covers 58 distinct accident categories. These videos were sourced from various existing public accident datasets, such as CCD~\cite{CCD}, A3D~\cite{A3D}, DoTA~\cite{DoTA}, and DADA-2000~\cite{Dada-2000}, in addition to streaming platforms like YouTube, Bilibili, and Tencent Video. It provides temporal labels for keyframes, including ``anomaly onset'', ``accident occur'', and ``accident end''. \textbf{Nexar}~\cite{nexar2025dashcamcollisionprediction} consists of 1,500 real-world dashcam video clips, each approximately 40 seconds long. Videos are recorded by Nexar dashcams with a resolution of 1280x720 at 30 frames per second. 
It was originally compiled for the Kaggle competition, Nexar Dashcam Crash Prediction Challenge. The test set is composed of 1344 videos with a duration of approximately 10 seconds, and we follow the preprocessing protocol strictly as ~\cite{nexar2025dashcamcollisionprediction}.

\textbf{Evaluation metrics.}
\label{sec:metrics}
We follow the evaluation protocol of Nexar~\cite{nexar2025dashcamcollisionprediction} and compute all metrics under a strict false alarm rate (FAR) constraint of $\lambda = 0.1$, as suggested by~\cite{zhao2025accident}. This setting ensures a realistic and fair comparison by maintaining a balanced trade-off between FAR and mean time-to-accident (mTTA). According to Nexar~\cite{nexar2025dashcamcollisionprediction}, we sample positive clips from four intervals before the accident: $[0.0, 0.5)$, $[0.5, 1.0)$, $[1.0, 1.5)$, and $[1.5, 2.0)$ seconds, with an equal number of negative clips from safe driving segments. For each interval starting at $\tau$, we compute:

\begin{itemize}
    \item \textbf{Constrained AUC$^\lambda_\tau$} and \textbf{AP$_\tau$}: According to ~\cite{zhao2025accident}, AUC and AP are calculated only when FAR $\leq 0.1$. 
    AUC$^\lambda_{0.0\mathrm{s}}$ reflects accident detection capability, while later intervals measure anticipation performance.
    \item \textbf{mTTA$^\lambda$}: Mean TTA of true positive alarms, computed only when FAR $\leq 0.1$. Note that mTTA requires the annotation of anomaly onset and is therefore used only as a compatibility metric for fair comparison with existing baselines, not for training. To avoid inflated values, we consider only alarms triggered \textit{after} anomaly onset. 
\end{itemize}

We report the primary metrics as:
\begin{align}
\mathrm{mAUC}^\lambda &= \frac{1}{3} \sum_{\tau \in \{0.5,1.0,1.5\}} \mathrm{AUC}^\lambda_\tau, \\
\mathrm{mAP} &= \frac{1}{3} \sum_{\tau \in \{0.5,1.0,1.5\}} \mathrm{AP}_\tau, \\
\mathrm{mTTA}^\lambda &= \mathbb{E}_{\theta:\,\mathrm{FAR}(\theta)\le \lambda}
\!\left[\mathrm{TTA}^{\theta}\right].
\end{align}

\begin{table*}[htbp]
\caption{Ablation study about different annotation strategies on the CAP~\cite{CAP} dataset, where FFR denotes Future-Frame Regularization, and AMC denotes Adaptive Monotonic Constraint.}
\centering
\footnotesize 
\renewcommand{\arraystretch}{1.4} 
\setlength{\tabcolsep}{3.5pt}     
\scalebox{0.96}{
\begin{tabular}{c|c|c|c|c|c|c||c|c|c|c||c|c|c|c}
\hline
\multirow{2}{*}{\textbf{Exp.}} & 
\multirow{2}{*}{\textbf{FFR}} & 
\multirow{2}{*}{\textbf{AMC}} & 
\multicolumn{4}{c||}{\textbf{Anomaly Onset Label}} &     
\multicolumn{4}{c||}{\textbf{Fixed Interval Label}} &     
\multicolumn{4}{c}{\textbf{Only Collision Label}} \\
\cline{4-15}
 & & & \textbf{mAUC$^{0.1}$} & \textbf{mAUC} & \textbf{mAP} & \textbf{mTTA$^{0.1}$(s)} & 
 \textbf{mAUC$^{0.1}$} & \textbf{mAUC} & \textbf{mAP} & \textbf{mTTA$^{0.1}$(s)} & 
 \textbf{mAUC$^{0.1}$} & \textbf{mAUC} & \textbf{mAP} & \textbf{mTTA$^{0.1}$(s)} \\
\hline \hline
I    &  &  & 0.441 & 0.829 & 0.871 & 1.162 & 0.412 & 0.816 & 0.872 & 1.173 & 0.358 & 0.783 & 0.832 & 1.009 \\
II   &  & \checkmark & 0.436 & 0.828 & 0.866 & 1.137 & 0.436 & 0.828 & 0.872 & 1.173 & 0.383 & 0.790 & 0.832 & 1.055 \\
III  & \checkmark &  & 0.444 & 0.830 & 0.872 & 1.174 & 0.455 & 0.847 & 0.886 & \textbf{1.212} & 0.474 & 0.850 & 0.850 & 1.202 \\
IV   & \checkmark & \checkmark & \textbf{0.484} & \textbf{0.856} & \textbf{0.887} & \textbf{1.198} & \textbf{0.480} & \textbf{0.851} & \textbf{0.890} & 1.203 & \textbf{0.483} & \textbf{0.853} & \textbf{0.890} & \textbf{1.207} \\
\hline
\end{tabular}
}
\label{tab:abl1}
\end{table*}

\begin{table*}[htbp]
\caption{Ablation study about different annotation strategies on the Nexar~\cite{nexar2025dashcamcollisionprediction} dataset, where FFR denotes Future-Frame Regularization, and AMC denotes Adaptive Monotonic Constraint.}
\centering
\footnotesize 
\renewcommand{\arraystretch}{1.4} 
\setlength{\tabcolsep}{3.5pt}     
\scalebox{0.96}{
\begin{tabular}{c|c|c|c|c|c|c||c|c|c|c||c|c|c|c}
\hline
\multirow{2}{*}{\textbf{Exp.}} & 
\multirow{2}{*}{\textbf{FFR}} & 
\multirow{2}{*}{\textbf{AMC}} & 
\multicolumn{4}{c||}{\textbf{Anomaly Onset Label}} &     
\multicolumn{4}{c||}{\textbf{Fixed Interval Label}} &     
\multicolumn{4}{c}{\textbf{Only Collision Label}} \\
\cline{4-15}
 & & & \textbf{mAUC$^{0.1}$} & \textbf{mAUC} & \textbf{mAP} & \textbf{mTTA$^{0.1}$(s)} & 
 \textbf{mAUC$^{0.1}$} & \textbf{mAUC} & \textbf{mAP} & \textbf{mTTA$^{0.1}$(s)} & 
 \textbf{mAUC$^{0.1}$} & \textbf{mAUC} & \textbf{mAP} & \textbf{mTTA$^{0.1}$(s)} \\
\hline \hline

I    &  &  & 0.388 & 0.826 & 0.822 & 0.773 & 0.420 & 0.837 & 0.827 & 0.741 & 0.298 & 0.789 & 0.781 & 0.610 \\
II   &  & \checkmark & 0.408 & 0.844 & 0.842 & 0.902 & 0.445 & 0.815 & 0.830 & 0.815 & 0.302 & 0.797 & 0.785 & 0.654 \\
III  & \checkmark &  & 0.434 & 0.846 &  0.850 & 0.897 & 0.434 & 0.865 & 0.856 & 0.920 & 0.453 & 0.847 & 0.854 & 0.836 \\
IV   & \checkmark & \checkmark & \textbf{0.479} & \textbf{0.872} & \textbf{0.876} & \textbf{0.951} & \textbf{0.454} & \textbf{0.875} & \textbf{0.874} & \textbf{0.931} & \textbf{0.472} & \textbf{0.869} & \textbf{0.870} & \textbf{0.958} \\
\hline
\end{tabular}
}
\label{tab:abl2}
\end{table*}

\textbf{Implementation details.}
In our experimental setup, training snippets are sampled only from the pre-collision period. During testing we apply a causal sliding window to the pre-collision portion of each accident video to obtain frame-level predictions.
We employ a 3D CNN as the snippet encoder to jointly extract spatial and temporal features. After encoding, we apply only adaptive spatial average pooling, preserving the temporal dimension of the feature sequence. 
In the transform pipeline, we resize each video frame to 224$\times$224 and resample frames to a frame rate of 10 frames per second, so that the model can process the frames within a 0.1s time interval.
We set the number of observed frames (the number of frames in a snippet) to 5. We use SlowOnly~\cite{slowfast} as our snippet encoder and train our model based on their pre-trained weights. We set the adaptive monotonic constraint sampling hyperparameters $d_{min}=0.1$, $d_{max}=0.9$, $\delta_0=0.01$. We set the regularization terms balancing hyperparameters $\lambda_1=1.5$, $\lambda_2=1.1$, and optimize the model with the SGD optimizer. Our model was trained for 50 epochs on 8 NVIDIA A800 GPUs with a batch size of 64, using an initial learning rate of 0.002 that was decayed to 10\% of its previous value every 20 epochs.

\subsection{Comparison with State-of-the-Art Methods}

We compare the quantitative results of our method with previous methods AdaLEA~\cite{Adaptive-Loss}, DSTA~\cite{karim2022dynamic}, GSC~\cite{wang2023gsc}, CAP~\cite{CAP}, XAI~\cite{XAI} and CRASH~\cite{CRASH} on both the CAP dataset and the Nexar dataset, to demonstrate its effectiveness. All baselines are re-trained following the evaluation protocol in Nexar~\cite{nexar2025dashcamcollisionprediction}, ensuring a fair comparison under the constrained false-alarm rate setting ($\lambda=0.1$). The results are shown in Table~\ref{tab:comparison}. 

On the CAP dataset, our method significantly outperforms previous approaches, achieving mAUC$^{0.1}$ score of 0.483. 
While achieving a top-tier mAP of 0.890, our model also delivers the earliest warnings, leading all compared methods with the best mean Time-to-Accident (mTTA$^{0.1}$) of 1.207 seconds. Notably, DSTA achieves the highest mAUC (0.895), but our method maintains competitive performance (0.853) while excelling in other critical metrics. 

The superiority of our approach is further emphasized on the more challenging Nexar dataset. Our method establishes state-of-the-art results across all metrics: mAUC$^{0.1}$ (0.472), mAUC (0.869), and mAP (0.870), which surpass the second best method CRASH by 0.079, 0.037 and 0.024, respectively. It also achieves the highest mTTA$^{0.1}$ of 0.958 seconds, underscoring its robust and effective capability for early accident anticipation. The above results demonstrate that our model not only predicts accidents with better accuracy but also provides more timely alerts, validating its effectiveness for real-world safety systems.

\subsection{Ablation Study}
We conduct an extensive ablation study on the CAP and Nexar datasets to evaluate the effectiveness of our proposed components: the future-frame regularization loss (FFR) and the adaptive monotonic constraint loss (AMC). Tab.~\ref{tab:abl1} and \ref{tab:abl2} report results on the two datasets under three different annotation paradigms: (1) Anomaly Onset, where frames from the manually annotated anomaly start to the collision are labeled as positive; (2) Fixed Interval, where a fixed 2-second window before the collision is labeled as positive; and (3) Only Collision, where only the collision frame is labeled as positive sample, while other frames are left unlabeled.

\textbf{Effectiveness of the future-frame regularization loss.} As shown across both tables, our FFR loss consistently improves performance across all labeling strategies and datasets. Specifically, comparing with Exp. I and Exp. III, we observe that applying only the FFR constraint provides the most significant performance gain. On CAP, under the three annotation schemes, it increases the mAUC$^{0.1}$ by 0.003, 0.043 and 0.116, respectively. On Nexar, under the three annotation schemes, it improves the mAUC$^{0.1}$ by 0.091, 0.034 and 0.174, respectively. This confirms that enforcing temporal correlation through future frame regularization effectively propagates the high-risk signal from the collision frame backward in time. 

We also observe that the baseline (Exp. I) of our Only Collision setting yields the worst performance among the three settings, where we obtain mAUC$^{0.1}$ of 0.358 on CAP and 0.298 on Nexar. However, by incorporating FFR (Exp. III), the performance is dramatically boosted to a mAUC$^{0.1}$ of 0.474 on CAP and 0.453 on Nexar—a remarkable improvement of 0.116 and 0.155, respectively. 

\begin{figure*}[t]
    \centering
    \includegraphics[width=\textwidth, keepaspectratio]{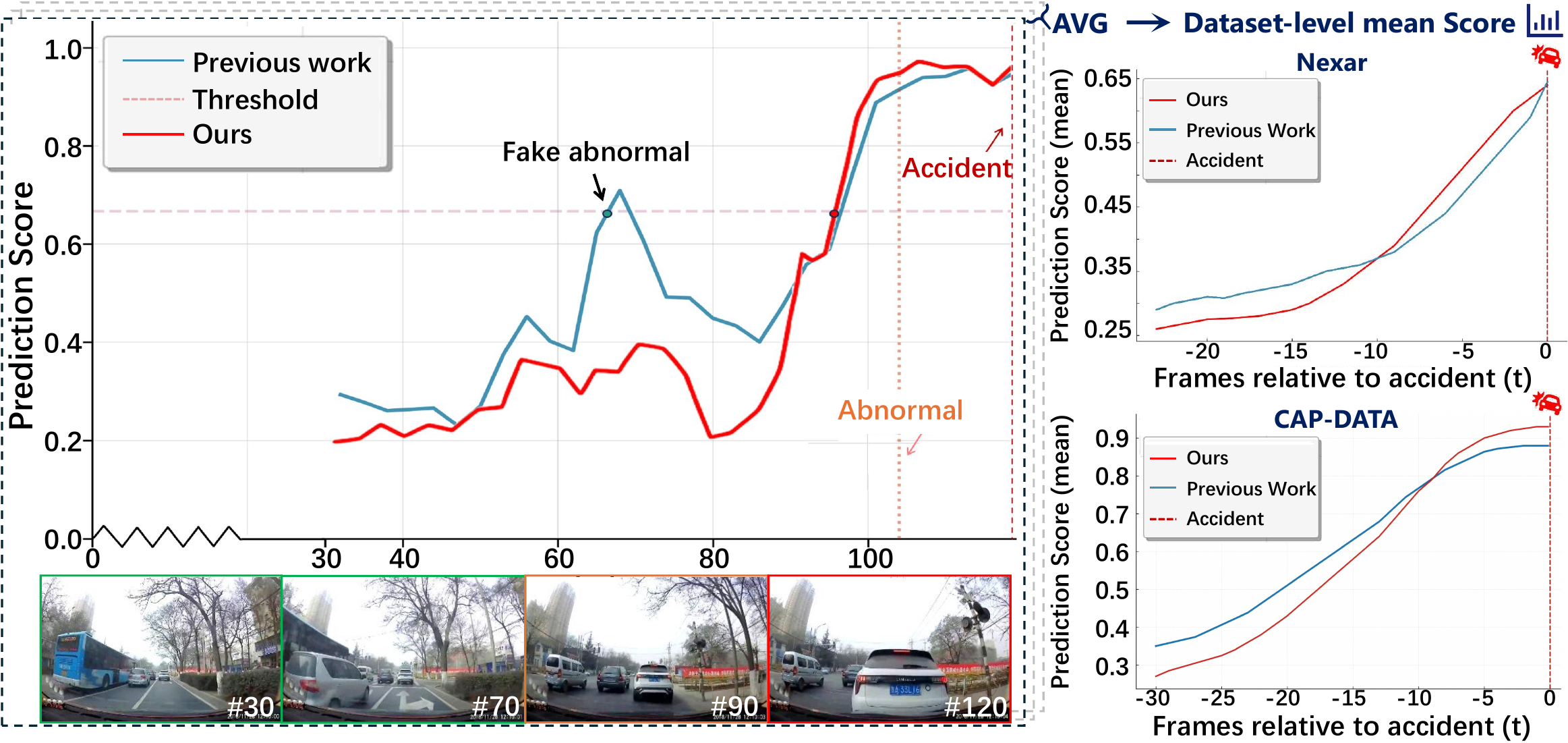}
    \caption{
        \textbf{Frame- and Dataset-Level Risk Prediction.}
Left: On a representative accident video, supervised baselines produce early false peaks before clear risk emerges. Our RiskProp model prediction remains low during safe periods and rises sharply only when discriminative cues appear, yielding a temporally coherent risk curve.
Right: Dataset-wide average risk curves on CAP and Nexar, aligned to accident timestamps. RiskProp suppresses early false positives and delivers a steeper, better-calibrated rise near the event, enabling earlier, more reliable warnings.
    }
    \label{fig:vis}
\end{figure*}

\textbf{Effectiveness of adaptive monotonic constraint loss.} As shown across both tables (Exp., I vs. Exp. IV), our AMC loss improves performance significantly across all labeling strategies and datasets when combined with FFR.
Specifically, on Nexar, under the three annotation schemes, it increases the mAUC$^{0.1}$ by 0.02, 0.025 and 0.004, respectively. The consistent improvements demonstrate that adaptive monotonic constraint loss enforces the risk score increase over time. This enhances the reliability of the predictive curve.

The combination of both constraints yields the best overall results. Adding AMC on top of FFR (Exp. IV) further refines the predictions, leading to the highest scores in nearly all metrics on both datasets, such as improving the mAP from 0.850 to 0.890 on CAP. This indicates that the monotonic risk trend enforced by AMC loss acts as a powerful regularizer for the temporally smooth predictions generated by the future-frame regularization loss.

\subsection{Analysis}

\textbf{Effectiveness of only-collision supervision with our self-supervised constraints.}
A key finding from our study is that our model, empowered by the FFR and AMC constraints, reduces the dependence on dense temporal annotations, though collision labels are still required. While the Only Collision strategy performs the worst without our constraints (Exp. I) due to a lack of full supervision, it achieves state-of-the-art results when they are applied (Exp. IV).

Remarkably, on both datasets, the full model trained with the Only Collision strategy (Exp. IV) performs on par with, and in some cases better than, the same model trained with more complex and costly labels. For instance, on the CAP dataset, its mAUC$^{0.1}$ of 0.483 is competitive with the 0.484 from the full Annotation setting. On the Nexar dataset, its mAP of 0.870 and mTTA$^{0.1}$ of 0.958s are the highest across all experiments. This is a significant result: it proves that our method can reach performance levels close to those achieved by dense annotation, without requiring humans to subjectively annotate the ``start'' of an accident. The proposed self-supervised constraints effectively compensate for the lack of dense supervision, enabling a more practical and scalable approach to accident anticipation.

In summary, our ablation studies validate two core conclusions: 1) FFR and AMC are essential components that significantly improve prediction accuracy. 2) When combined with these constraints, our Only Collision supervision matches the performance of densely labeled approaches, offering a powerful yet annotation-efficient solution.

\textbf{Analysis of the predicted risk curve.} As shown in Fig.~\ref{fig:vis}, our method produces more realistic and reliable risk curves than prior work. When a vehicle appears on the side at frame 70, the baseline method falsely raises its risk score above the alert threshold, triggering a premature false alarm. In contrast, our method remains a low risk estimate during this safe period, only exhibiting a sharp, sustained rise when the leading vehicle becomes critically close around frame 90, precisely when real danger appears. This shows that our self-supervised framework, guided by future-frame regularization and adaptive monotonicity constraints, effectively suppresses false positives induced by ambiguous cues, ensuring alerts are triggered only upon clear evidence of a risky accident. Consequently, the resulting risk evolution is smooth, temporally coherent, and aligns with the physical progression of real-world driving accidents.

\section{Conclusion}
In this work, we propose RiskProp, a novel collision-anchored self-supervised risk propagation paradigm for modeling risk evolution in early accident anticipation. Instead of learning from binary labels that indicate whether a frame is risky, our method uses the model’s own prediction at the next frame as a soft supervision signal for the current frame. Therefore, the risk information anchored at the collision point can be gradually propagated backward through the sequence. This shifts the training paradigm from static risk classification to dynamic risk evolution modeling, enabling a more temporally aware and physically plausible anticipation process.
To further shape this evolution, we incorporate an adaptive monotonic constraint loss, which encourages a generally increasing long-term risk trend as the scene progresses toward an accident, while still allowing natural short-term fluctuations. Together, these losses guide the model to learn smooth, irreversible risk evolution without requiring subjective anomaly-onset labels.
Extensive experiments on CAP and Nexar show that RiskProp achieves state-of-the-art performance with significantly more stable and interpretable risk curves.

\section*{Acknowledgments}
This work was supported by the National Natural Science Foundation of China (Grant No.~62471344), the Zhongguancun Academy (Project No.~20240304), and the CCF-DiDi GAIA Collaborative Research Funds for Young Scholars.
{
    \small
    \bibliographystyle{ieeenat_fullname}
    \bibliography{main}
}


\end{document}